\documentclass[10pt,twocolumn,letterpaper]{article}

\usepackage{cvpr}
\usepackage{graphicx}
\usepackage{amsmath}
\usepackage{amssymb}
\usepackage{booktabs}
\usepackage{multirow}
\usepackage{makecell}
\usepackage{array}
\usepackage{xcolor}
\usepackage{tikz}
\usetikzlibrary{arrows.meta,positioning,fit,calc}
\definecolor{cvprblue}{rgb}{0.21,0.49,0.74}
\usepackage[pagebackref,breaklinks,colorlinks,allcolors=cvprblue]{hyperref}
\usepackage{bbm}

\newcommand{\cls}[1]{\textsc{#1}}

\newif\ifcomments
\commentstrue
\ifcomments
    \providecommand\rotem[1]{\textcolor{red}{[Rotem: #1]}}
    \providecommand\ilan[1]{\textcolor{blue}{Ilan: #1]}}
    \providecommand\amit[1]{\textcolor{magenta}{[Amit: #1]}}
    \providecommand\todo[1]{\textcolor{orange}{[TODO: #1]}}
\else
    \providecommand{\rotem}[1]{}
    \providecommand{\ilan}[1]{}
    \providecommand{\amit}[1]{}
    \providecommand{\todo}[1]{}
\fi

\begin{document}

\title{Online Mistake Detection in Instruction-Conditioned Procedural Video}


\author{Amit Feldman\\
\and Ilan Shimshoni\\ 
{\tt\small ishimshoni@is.haifa.ac.il}
\and Rotem Dror\\
{\tt\small rdror@is.haifa.ac.il}\\
Faculty of computer and Information Science, University of Haifa\\
}



\maketitle

\begin{abstract}
An assistant monitoring a person performing a procedure is most useful when it can detect mistakes while there is still time to recover. This requires judging actions against task instructions as the video unfolds, identifying the attempted step, locating it in time, and assessing its execution. Existing online detectors do not use the instruction text, while instruction-conditioned methods operate offline. In this paper, we study online, instruction-conditioned mistake detection in egocentric videos. Using only observed video, task instructions, and prior predictions, our model identifies and localizes completed action attempts and classifies them as correct, modified, slip, corrections, or additions; omitted steps are determined after the video ends. We compare a modular pipeline, with separate step identification, temporal localization, and execution verification, to a joint model that predicts all outputs in a single step. To support supervised learning from mistakes, we repartition EgoPER to include erroneous videos in all splits. The approaches achieve similar step-segmentation F1@0.5 (93.6\% and 92.1\%), while the joint model achieves higher binary mistake-detection F1@0.5 (84.1\% vs.\ 80.7\%), with its largest gain on slips (56.4\% vs.\ 37.5\%). Component-level analysis suggests that error propagation between stages contributes to the modular approach's lower detection score. On test videos containing mistakes, both designs outperform released AEM and ESTANet checkpoints under our evaluation protocol.
\end{abstract}

\section{Introduction}
\label{sec:intro}
\definecolor{modblue}{RGB}{0,114,178}
\definecolor{jointorange}{RGB}{230,159,0}
\definecolor{rejectred}{RGB}{213,94,0}
\definecolor{iogray}{RGB}{120,120,120}
\providecommand{\cls}[1]{\textsc{#1}}
\begin{figure*}[ht]
\centering
\begin{tikzpicture}[
  font=\footnotesize,
  >={Latex[length=1.8mm,width=1.5mm]},
  block/.style={draw=#1!80!black, fill=#1!12, rounded corners=2pt,
    line width=0.6pt, align=center, inner sep=3pt},
  module/.style={block=modblue, text width=25.5mm, minimum height=13mm},
  flow/.style={->, line width=0.7pt, draw=black!80},
  reject/.style={->, line width=0.6pt, densely dashed, draw=rejectred},
  lbl/.style={font=\scriptsize, inner sep=1.5pt},
  rlbl/.style={lbl, text=rejectred!80!black, yshift=-1pt},
  panel/.style={rounded corners=4pt, line width=0.5pt},
  ptitle/.style={font=\footnotesize\bfseries, anchor=north west,
    inner sep=4pt},
]
\draw[panel, draw=modblue!55, fill=modblue!4]
  (2.8,-1.55) rectangle (13.3,1.45);
\draw[panel, draw=jointorange!75, fill=jointorange!6]
  (2.8,-4.45) rectangle (13.3,-1.8);
\node[ptitle, text=modblue!70!black] at (2.8,1.45)
  {(a) Modular approach};
\node[ptitle, text=jointorange!60!black] at (2.8,-1.8)
  {(b) Joint approach};

\draw[block=iogray] (0,-4.45) rectangle (2.5,1.45);
\node[font=\footnotesize\bfseries] at (1.25,-0.4) {Streaming input};
\foreach \i in {0,...,4}
  \draw[fill=iogray!35, draw=iogray!90!black, line width=0.4pt]
    ({0.33+\i*0.38},-1.13) rectangle ++(0.32,0.26);
\node[lbl] at (1.25,-1.43) {window $V[\tau,t]$, 1\,fps};
\node[lbl, align=center] at (1.25,-2.4)
  {task description $d$\\[1pt] instruction list $I$\\[1pt]
   prediction history};

\draw[block=iogray] (13.6,-4.45) rectangle (16.1,1.45);
\node[font=\footnotesize\bfseries] at (14.85,-0.4) {Accepted record};
\node[lbl, align=center] at (14.85,-1.15)
  {step $s$, interval $[a,b]$,\\[1pt] class $c$ + rationale};
\draw[iogray!70, line width=0.4pt] (13.95,-1.75) -- (15.75,-1.75);
\node[lbl, align=center] at (14.85,-2.4)
  {update history\\[1pt] $\tau \leftarrow \tau + b$\\[1pt]
   $t \leftarrow \tau + \Delta$};

\node[module] (sid) at (4.45,0)
  {\textbf{Step Identifier}\\[1pt]\scriptsize which instruction\\
   was attempted?};
\node[module] (loc) at (8.05,0)
  {\textbf{Temporal Localizer}\\[1pt]\scriptsize when did the\\
   attempt start and end?};
\node[module] (ver) at (11.65,0)
  {\textbf{Execution Verifier}\\[1pt]\scriptsize was it executed as\\
   instructed? (on crop)};

\draw[flow] (2.5,0) -- (sid.west);
\draw[flow] (sid) -- node[lbl, above] {$s$} (loc);
\draw[flow] (loc) -- node[lbl, above] {$[a,b]$} (ver);
\draw[flow] (ver.east) -- node[lbl, above] {$c$} (13.6,0);

\draw[flow] (loc.north) |-
  node[lbl, above, pos=0.73]
  {$s=\texttt{EXTRA}$: label \texttt{ADDITION}, skip verification}
  (13.6,1.0);

\foreach \n in {sid,loc,ver}
  \draw[reject, -] (\n.south) -- (\n.south |- 0,-1.0);
\foreach \n in {sid,loc}
  \fill[rejectred] (\n.south |- 0,-1.0) circle (0.9pt);
\draw[reject] (ver.south |- 0,-1.0) -- (2.5,-1.0);
\node[rlbl, below] at (7.25,-1.0)
  {\texttt{NOT\_COMPLETED} from any module: keep $\tau$,
   extend $t \leftarrow t + \Delta$};

\node[block=jointorange, text width=97mm, minimum height=12mm]
  (joint) at (8.05,-2.95)
  {\textbf{Joint model}: step identity, boundaries, and execution
   class in one response\\[2pt]
   \scriptsize\texttt{\{step\_id, start, end, classification,
   reasoning\}}};
\draw[flow] (2.5,-2.95) -- (joint.west);
\draw[flow] (joint.east) -- (13.6,-2.95);
\draw[reject] (joint.south) -- (joint.south |- 0,-3.9) -- (2.5,-3.9);
\node[rlbl, below] at (5.3,-3.9)
  {\cls{Not\_completed}: keep $\tau$, extend $t \leftarrow t + \Delta$};

\draw[flow] (14.85,1.45) -- (14.85,2.0) --
  node[lbl, above]
  {accepted: next window starts at the predicted end of the action}
  (1.25,2.0) -- (1.25,1.45);
\end{tikzpicture}
\caption{%
Overview of the two approaches. At each call, the system receives the
current window $V[\tau,t]$ (sampled at 1\,fps), the task description
$d$, the instruction list $I$, and its prediction history.
\textbf{(a)}~The modular approach chains a Step Identifier, which
selects an instruction $s\in I$ or \texttt{EXTRA}; a Temporal Localizer,
which predicts the attempt's boundaries $[a,b]$; and an Execution
Verifier, which classifies the cropped attempt
$V[\tau{+}a,\,\tau{+}b]$ as \texttt{CORRECT}, \texttt{MODIFICATION},
\texttt{SLIP}, or \texttt{CORRECTION}, using its previous execution labels as
context. Extra actions are labeled \texttt{ADDITION} without verification.
\textbf{(b)}~The joint approach predicts all fields in a single
structured response, conditioned on up to 12 previous predictions.
In both, a \texttt{NOT\_COMPLETED} output (dashed) keeps the window start
$\tau$ and extends the window by $\Delta = 5$\,s; an accepted record
updates the history and starts the next window at the predicted end.
}
\label{fig:overview}
\end{figure*}
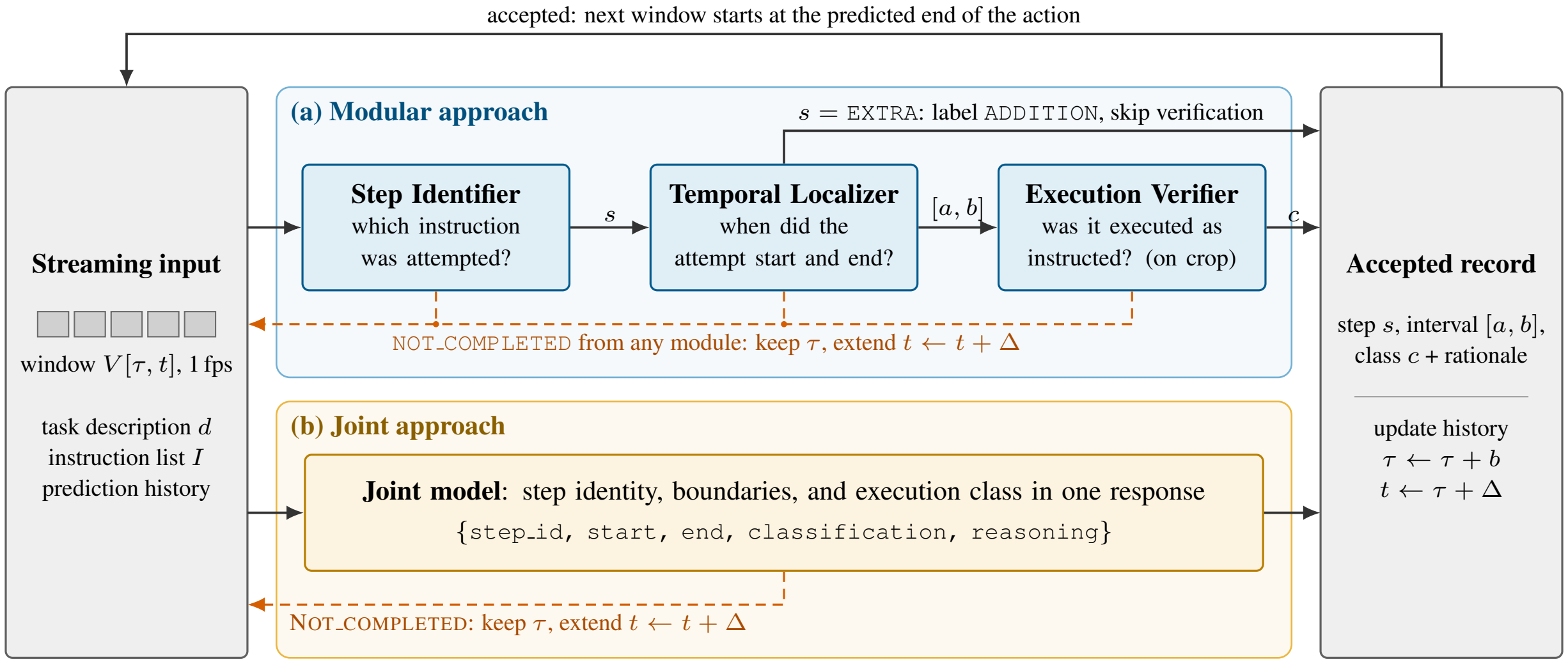

Procedural mistake detection asks whether and how a person performing a multi-step task, such as cooking, assembling furniture, or following laboratory protocols, deviates from the prescribed procedure, as defined by its instructions.
In computer vision, this is studied as a video-understanding
problem: given egocentric video of a person carrying out a
task, the system must recognize and temporally localize actions
and determine whether they adhere to the procedure's instructions
\cite{lee2024egoper,flaborea2024prego}. This capability could
support an assistant that monitors a person recording themselves
while performing a multi-step procedure, and notify them if a mistake has been made while they can still correct it. To provide such feedback, a model must interpret the video as it unfolds; since feedback delivered only after the task has failed may explain what went wrong but cannot help the person recover when the error occurred.

This problem is challenging because visual appearance alone
does not determine whether an action is correct. Pouring water
into a bowl is an ordinary action, but it is a mistake if the
instructions specify pouring the water into the bowl containing oats. The model must therefore relate visual evidence to the task instructions and distinguish between different deviations: omitting or adding a step, performing an instructed step differently, failing to achieve its goal, or taking an action that repairs an earlier
failure. These distinctions determine what went wrong and
whether the procedure can still be completed successfully.

Prior work has addressed parts of this problem in both
offline and online settings. Offline methods use the complete
recording to segment the video or align it with the procedure's steps
before evaluating the actions~\cite{lee2024egoper,
huang2025amnar,lee2025gtg2vid}. Instruction-conditioned
approaches, including EgoOops~\cite{haneji2024egooops} and
ZeProM~\cite{ozsoy2026zeprom}, compare observed actions
directly with procedural text, but make their judgments
after observing the entire recording. 
Online methods make decisions from
the video observed so far. PREGO~\cite{flaborea2024prego}
detects deviations in action order, while ESTANet
\cite{lee2026estanet} also detects execution errors.
However, these online methods do not assess actions against
the explicit task instructions and do not provide the same
combination of attempted-step identity, temporal boundaries,
and fine-grained execution labels. This leaves a gap between
online monitoring and instruction-conditioned assessment:
an online assistant needs to determine not only that an
action is incorrect, but which instruction it attempted
and how its execution differs from that instruction.

We address this gap by studying online, instruction-conditioned
mistake detection in egocentric video. At each decision, the
system uses the video observed so far, the task
description and instruction list, and its previous predictions.
It identifies a completed action attempt, localizes it in
time, and classifies its execution as correct, a modification,
a slip, a correction, or an addition. Because an omitted step
has no observed action interval, omissions are identified
separately from the actions detected during the stream and after the entire video has been processed.
We compare two prediction designs: a modular approach that
uses a Step Identifier, Temporal Localizer, and Execution
Verifier in sequence, and a joint approach that predicts the
same outputs in one response. Both maintain prediction
history so that later actions, such as a correction, can be
interpreted in the context of earlier events. An illustration of both methodologies appears in~\cref{fig:overview}.

We evaluate our approaches on EgoPER~\cite{lee2024egoper},
a dataset of egocentric videos of five cooking tasks, with
annotations for procedural steps and mistakes.
Because the official training and validation sets contain
only normal videos (without mistakes), we construct a task- and error-stratified split that includes erroneous examples in training and
validation, while retaining both normal and erroneous videos
for testing. This enables us to study supervised, online
mistake detection across the full evaluation set and separately
on videos containing errors.

Our results reveal a distinction between recognizing the
procedure and detecting its mistakes. The modular and joint
approaches achieve similar temporal step-segmentation
performance, but the joint approach attains higher error
detection F1. A modular component analysis helps explain this pattern: localization and execution classification perform well when their upstream predictions are correct, yet errors in those predictions can propagate through the pipeline. Predicting step identity, boundaries, and execution together reduces this dependency and improves end-to-end mistake detection. Thus, our findings suggest that the main benefit of the joint design lies in integrating
decisions about whether an action is erroneous and how it
relates to the instructions, rather than in improving temporal
segmentation alone.
In addition, both suggested approaches outperform the evaluated AEM and ESTANet checkpoints on the erroneous-video test subset, under our rigorous evaluation protocol.

The rest of the paper is organized as follows. We first review
related work on online and instruction-conditioned mistake
detection. We then define the task and describe the modular
and joint approaches. Next, we present the dataset split,
evaluation metrics, and implementation details, followed by
the main results and analyses of fine-grained mistake
classification, temporal segmentation, and component-level
performance.


\section{Related Work}
\label{sec:related}

Mistake detection in procedural videos is studied in two
settings. \emph{Offline} methods access the complete recording
before evaluating an action, whereas \emph{online} methods
make decisions as the video unfolds, using only observations
available so far~\cite{flaborea2024prego,lee2026estanet}. 
In the online setting, the literature also defines \emph{real-time} mistake detection, which has strict requirement of instantaneous response.
Since our work addresses online mistake
detection, we focus our review on methods operating in this
setting. We also review instruction-conditioned methods,
which assess observed actions against explicit procedural
instructions. This perspective is central to our formulation:
a mistake is a deviation from the provided instructions,
rather than from the model's general expectations about
how the task should be performed.

In this setting, the literature commonly separates \emph{procedural} mistakes (omitted, added or reordered steps) from \emph{execution} mistakes (a step performed with the wrong technique, tool, quantity or timing)~\cite{bacharidis2025review,haneji2024egooops}. We use these terms to describe the types of mistakes recent work aims to detect.

\subsection{Online Mistake Detection}
Existing approaches to online mistake detection differ
in whether they detect deviations in the action sequence,
errors in execution, or both.
PREGO~\cite{flaborea2024prego} detects procedural mistakes
by comparing recognized actions with anticipated ones.
Its recognition branch identifies ongoing actions, while
its anticipation branch prompts a large language model (LLM) with previously recognized actions and examples of correct sequences.
A mismatch between the recognized and anticipated step
signals a mistake. Because this comparison operates on
step labels, an incorrectly executed action can escape
detection if it is recognized as the expected step.
PREGO also introduced Assembly101-O and Epic-Tent-O,
adaptations of the original datasets~\cite{sener2022assembly101,
jang2019epic} in which test videos end at the first
procedural error. These benchmarks therefore do not
evaluate continued monitoring after a mistake or a recovery.

More recent methods also examine execution quality.
ESTANet~\cite{lee2026estanet} detects errors through
inconsistencies among action predictions for the same
video clip. Standard and error-sensitive detectors
respond differently to execution errors, while detectors
with different temporal contexts expose deviations
from the intended action order. This allows ESTANet
to detect both execution and procedural errors.
However, it does not condition its decisions on
instruction text, distinguish fine-grained mistake
types such as slips and modifications, or explicitly
associate an error with the instruction attempted
and the required execution.

Beyond detecting individual mistakes,
PROVIA~\cite{wen2026provia} emphasizes continued
procedure tracking through errors and recoveries.
Our work likewise maintains monitoring after recovery,
however, we also ground predictions in explicit instructions
and identify the attempted step and its execution
class. This motivates our review of instruction-conditioned
methods, which assess actions against procedural text
but have largely been studied in the offline setting.

\subsection{Instruction-Conditioned Mistake Detection}
Offline mistake detectors use the complete video to segment
the procedure into steps and evaluate each step using context
from the full recording~\cite{lee2024egoper,huang2025amnar,lee2025gtg2vid}.
Among these, \emph{instruction-conditioned} methods assess
each step against textual instructions specifying how the
procedure should be performed. This allows them to identify
actions that appear reasonable in isolation but constitute
mistakes because they violate an explicit requirement,
such as using tape when the manual specifies glue.

\emph{EgoOops} is an example of this approach. Introduced by
\citet{haneji2024egooops}, it pairs
egocentric videos of five tasks with procedural text,
temporal step alignments, a mistake taxonomy, and
justifications. Their detection method aligns the video
with the instructions, then uses a multilayer perceptron network (MLP) to classify
each aligned clip as a correct execution, a mistake,
or a correction.

More recently, vision-language models (VLMs) have enabled direct comparison between observed actions and instruction text.
ZeProM~\cite{ozsoy2026zeprom} prompts a single pre-trained
VLM with the complete video and an ordered list of
recipe steps, without additional training. The model
segments the video, associates each segment with a step,
identifies errors with brief explanations, and lists
omitted steps.
Both methods assess instruction adherence after observing
the complete recording, limiting their usefulness for
timely intervention. Our models perform this task
online, detecting mistakes while the procedure is underway
and corrective action may still be possible.

\section{Methodology}
\label{sec:method}

Given a procedural task description, its instructions in natural language, and an ongoing egocentric video stream of a user performing the task, we propose detecting instruction deviations in an online manner by performing three steps: identifying completed instruction-step attempts, localizing them in time, and determining whether their execution follows the instructions correctly.
We propose a modular approach, which separates these decisions into three components, and a joint approach, which predicts them in a single response, and compare between the performance of the two approaches.

\subsection{Problem Setting}
\label{sec:problem-setting}

A procedural task is specified by a natural-language description $d$ and a task dependent ordered list of $K_d$ natural-language instructions $I=(i_1,\ldots,{i_K}_d)$. In this setting, the model observes an egocentric video $V$ incrementally, using only the available video and its earlier predictions.

We distinguish an instruction from an \emph{attempt}
to perform it. An attempt corresponds to a listed
instruction even when its execution is incorrect.
For example, pouring water into an empty bowl is
still an attempt at the instruction to pour water
into the bowl containing oats. A completed attempt
therefore means that the action has ended, rather
than that its intended goal has been achieved.

For each detected attempt, the system predicts a step
identity $s$, temporal boundaries $[a,b]$, and an
execution class $c$. The step identity
$s\in\{1,\ldots,K\}\cup\{\texttt{EXTRA}\}$ associates
the attempted action with a listed instruction or identifies
it as an additional action outside the procedure. 
The execution class takes one of the following labels:
\begin{description}
    \item[\texttt{CORRECT}]
    The action follows the specified instruction.
    \item[\texttt{MODIFICATION}]
    The step is performed differently from the instruction,
    for example by stirring with a knife instead of
    a spoon or sweetening tea with sugar instead of
    honey. Such a deviation does not necessarily
    change the outcome of the step.
    \item[\texttt{SLIP}]
    The execution fails to achieve the goal of the
    step and requires a subsequent corrective action.
    Examples include pouring water into a bowl other
    than the one containing oats. A slip is labeled as such whether
    or not the person subsequently corrects it.
    \item[\texttt{CORRECTION}]
    The action mitigates the effect of an earlier
    slip, for example by transferring misplaced
    water into the bowl containing oats.
    This label describes a recovery action rather
    than a new mistake.
    \item[\texttt{ADDITION}]
    The person performs an unnecessary step
    outside the specified procedure, such as adding
    raisins when the original recipe does not
    include them. In this case, $s=\texttt{EXTRA}$.
\end{description}

The distinction between modification and addition
depends on whether the action is an alternative
execution of a listed step or an extra step outside
the procedure. For example, replacing honey with
sugar while sweetening tea modifies an instructed
step, whereas adding an unrequested ingredient as
an extra action is considered an addition.

In the case where no completed attempt is identified, the system
returns \texttt{NOT\_COMPLETED}. This is not a classification label, but an indication that the temporal boundaries of the window returned by the model need to be extended.
Meanwhile, step omissions ($c=\texttt{OMISSION}$) are a separate output, defined in~\cref{sec:omissions}, because they do not correspond to observed action intervals.

To determine the temporal boundaries of actions, the video stream is processed continuously and sampled at one frame per second. The model applies a procedure for incremental window construction defined next. 

\subsection{Incremental Video Processing}
\label{sec:stream-processing}
Let $\tau$ denote the predicted end of the last attempted action ($\tau=0$ before any action has been predicted) and $t$ the end time of the constructed window. The model initially examines a five second window starting at $\tau$, i.e., $V[\tau, t=\tau+5]$, as the potential window in which an action should be completed.
If the model returns \texttt{NOT\_COMPLETED}, $\tau$ remains fixed and the endpoint, $t$, is extended by $\Delta=5$ seconds, i.e., $V[\tau, t+\Delta]$.
Thus, successive unsuccessful evaluations examine extending windows of 5, 10, 15 seconds, and so forth, all beginning at the same point.

When an action is accepted and identified, its predicted boundaries
$[a,b]$ are expressed relative to the current window
and satisfy $0\leq a<b\leq t-\tau$.
The corresponding absolute interval is
$[\tau+a,\tau+b]$. The next window starts at the
predicted action end and again has an initial
duration of five seconds:
    $\tau_{\mathrm{next}}=\tau+b, \quad
    t_{\mathrm{next}}=\tau_{\mathrm{next}}+\Delta$.
All windows are clipped to the video duration.
For example, if an action is detected in a window 
covering seconds 0--15 and its predicted end is
second 12, the next window covers seconds 12--17.
The system therefore resumes from the predicted
action boundary, retaining any subsequent activity
that appeared in the previous window.


\subsection{Modular Approach}
\label{sec:modular}
As mentioned above, we propose two approaches for solving the problem of online mistake detection. The two approaches are illustrated in~\cref{fig:overview}.
The modular approach consists of the following modules:
\paragraph{Step Identifier}
The step identifier receives the video in the current window
$V[\tau,t]$, the task description $d$, the complete
instruction list $I$, and a history of up to 12 previous eligible
action predictions.\footnote{This history length provides context for linking corrections to preceding slips while limiting the input length.}
It selects a listed instruction, \texttt{EXTRA}, or
\texttt{NOT\_COMPLETED}. Naturally, the model prefers the next step, but the identifier is not constrained to select instructions in their listed order.

We formulate this problem as multi-class classification. Hence, the output is the highest logit value label, rather than a freely generated response. 
If the prediction is \texttt{NOT\_COMPLETED}, the model extends the window and starts over (this can happen in every module). Otherwise, it passes the selected label to the Temporal Localizer. 


\paragraph{Temporal Localizer}
The temporal localizer receives the same video
window and the selected instruction, or the
\texttt{EXTRA} designation. It outputs the
start ($a$) and end ($b$) times of the corresponding attempt, even if the action was performed incorrectly.

The localizer can reject an incomplete attempt
with \texttt{NOT\_COMPLETED}. A rejection leaves
the window start unchanged and causes the
system to accumulate more video.
For an accepted \texttt{EXTRA} action, processing
ends at this component: the system assigns
\texttt{ADDITION} and advances to the next
window without invoking the execution verifier.
Other localized attempts proceed to verification.

\paragraph{Execution Verifier}
The execution verifier receives the localized
crop $V[\tau+a,\tau+b]$, the selected instruction,
definitions of the execution classes, and its
previous execution predictions. It classifies
the attempt as \texttt{CORRECT},
\texttt{MODIFICATION}, \texttt{SLIP}, or
\texttt{CORRECTION}, with a textual explanation.
If it rejects the crop as incomplete, the
accumulation window is extended without
advancing its start.

The execution history provides context for
interpreting the current action. In particular,
a correction may be recognizable only in
relation to a preceding slip that is no
longer visible in the localized crop.
This context is distinct from the history
of step identities supplied to the step
identifier. After verification, the system combines the
predicted step identity, interval, and execution
label into an accepted record, updates the
relevant prediction histories, and resets
the window according to~\cref{sec:stream-processing}.

\subsection{Joint Approach}
\label{sec:joint}

The joint approach uses a single adapted
vision-language model to identify, localize,
and classify an action in one response.
It receives the accumulated video $V[\tau,t]$,
the task description $d$, the instruction list
$I$, and a history of up to 12 previous eligible
action predictions. 

The output is either \texttt{NOT\_COMPLETED}
or a JSON record containing the step identity,
start and end times, execution class, and a
textual explanation of the detected deviation,
when applicable. 
As before, a \texttt{NOT\_COMPLETED} response extends the
current window by five seconds. An accepted
record updates the prediction history and
starts a new five-second window at the
predicted action end.

Joint prediction may outperform the modular
approach by avoiding error propagation between
components. In the modular approach, an
incorrect step identity affects localization
and verification, while an inaccurate interval
can remove evidence needed to classify the
execution. The joint model predicts these
outputs together, allowing their relationships
to inform a single decision. It still remains
susceptible to errors propagated across time
through its history and window updates. 

\subsection{Omission Detection}
\label{sec:omissions}

An omission occurs when one or more required
steps are skipped, such as not checking the
water temperature. It has no observed temporal
interval and is therefore handled separately
from action classification.
Let $\widehat{S}$ be the set of instruction
identities detected over the complete video.
The predicted omission set is $\widehat{O}=\{1,\ldots,K_d\}\setminus\widehat{S}$.
This set is computed from the full detection
record, rather than the bounded history used
as model input. A detected attempt counts as
observed even if its execution is incorrect.

During the procedure, the instruction order
can indicate that an expected step has not
yet been observed. Such an alert can only be
provisional: a step performed later is not
considered omitted. We therefore finalize
omissions only at the end of the video.

\section{Experimental Setup}
\label{sec:experimental-setup}

Our experiments compare the modular and joint approaches
under the same data split and video-processing procedure.
We evaluate whether each model detects mistakes, identifies
their types, associates them with the correct action
intervals, and assess step recognition and omission detection.

\subsection{Dataset and Evaluation Splits}
\label{sec:data}

We conduct our experiments on EgoPER~\cite{lee2024egoper}, which contains 389 egocentric videos of five cooking tasks: making coffee, oatmeal, pinwheels, quesadillas, and tea.
216 videos contain no mistakes and 173 contain
mistakes or corrective actions. The dataset provides temporal
step annotations for 4,741 segments: 4,237 correct executions,
215 modifications, 143 slips, 104 additions, and 42 corrections.

The official training and validation sets contain only normal
videos, while all erroneous videos belong to the test set.
Since our supervised models require examples of mistakes during training, we repartition the
videos into training, validation, and test sets, stratified
by task and error class, without changing their annotations.
Specifically, we move 148 of the 173 erroneous videos
(approximately 85\%) from the official test set into training
and validation, retaining 25 erroneous videos for testing. Normal videos are
also redistributed, with 183 assigned to training and validation
and 33 to testing. The resulting partition appears in~\cref{tab:egoper-repartition}.

\begin{table}[ht]
\centering
\small
\setlength{\tabcolsep}{5pt}
\caption{Repartitioning EgoPER for supervised mistake
detection.}
\begin{tabular}{llrrr}
\toprule
Partition & Subset & Normal & Erroneous & Total \\
\midrule
Official & Train + validation & 201 & 0   & 201 \\
         & Test               & 15  & 173 & 188 \\
\midrule
Ours     & Train + validation & 183 & 148 & 331 \\
         & Test               & 33  & 25  & 58  \\
\bottomrule
\end{tabular}
\label{tab:egoper-repartition}
\end{table}

\subsection{Baselines}
\label{sec:baselines}

Our main comparison is between the modular and joint approaches. For comparisons with existing mistake detectors, we evaluate released checkpoints of AEM~\cite{guo2026aem}, an offline mistake detection method, and ESTANet~\cite{lee2026estanet}, an online method, on the shared 25-video test subset that contains mistakes, because the other 33 test videos may belong to the training or validation sets on which they were trained. Both baselines were trained on correct videos, whereas our models also receive annotated mistakes during training. These comparisons therefore measure performance under different supervision regimes.

We evaluate AEM and ESTANet with our own evaluation code.
For AEM, we use its default mistake probability threshold, $\tau=0.5$.
For ESTANet, we group frames into segments by its predicted step label and mark a segment as a mistake when more than half of its frames are predicted as such.
We also evaluate predictions using ESTANet's released evaluator, which measures frame accuracy and segmental $F1@10/25/50$ (see supplementary).



\subsection{Evaluation Metrics}
\label{sec:metrics}
We report each metric as an unweighted mean across
the five recipes. For binary evaluation, modifications,
slips, additions, and corrections are grouped as errors,
while correct executions are labeled normal.
Corrections are included for scoring purposes,
although they represent recovery rather than new mistakes.

\paragraph{Error Detection Accuracy (EDA).}
An annotated segment is predicted as erroneous when
predicted error intervals cover strictly more than
half its duration. Let $E$ denote the union of these
intervals, $g_i$ an annotated segment, and $z_i$
its binary error label. Then
\begin{equation}
    \hat z_i=\mathbbm{1}\left[
        \frac{|g_i\cap E|}{|g_i|}>\frac{1}{2}
    \right],
    \quad
    \mathrm{EDA}=\frac{1}{N}\sum_{i=1}^{N}
        \mathbbm{1}[\hat z_i=z_i],
\end{equation}
where $N$ is the number of annotated segments.
EDA measures classification accuracy, but can be
dominated by the more frequent normal segments.

\paragraph{Binary F1@0.5.}
Correct segments substantially outnumber erroneous ones in
our test set (621 versus 73). As a result, accuracy alone can
remain high even when a method misses many errors. We
therefore report binary F1, which gives equal weight to the
error and normal classes by averaging their class-specific F1
scores. A method that labels every segment as normal, for
example, receives no error-class true positives and is
penalized accordingly. The metric also accounts for temporal
localization by comparing only predicted and annotated segments that their binary labels agree and their temporal intersection-over-union satisfies:
\begin{equation}
    \operatorname{tIoU}(p,g)
    =\frac{|p\cap g|}{|p\cup g|}
    \geq 0.5.
\end{equation}
For each class $c\in\{\mathrm{error},\mathrm{normal}\}$,
matched pairs count as true positives, unmatched
predictions as false positives, and unmatched
annotations as false negatives. We compute:
\begin{equation}
    F_{1,c}=
    \frac{2\mathrm{TP}_c}
    {2\mathrm{TP}_c+\mathrm{FP}_c+\mathrm{FN}_c},
\end{equation}
\begin{equation}
    F_{1,\mathrm{binary}}=
    \frac{F_{1,\mathrm{error}}+F_{1,\mathrm{normal}}}{2}.
\end{equation}

\paragraph{Omission Accuracy and IoU}
For a video, let $O$ be the set of annotated omitted steps and $\widehat{O}$ the set of recipe steps that the model never predicts. 
\begin{equation}
\text{O-Acc}=\frac{|\widehat{O}\cap O|}{|O|}, \quad \text{O-IoU}=\frac{|\widehat{O}\cap O|}{|\widehat{O}\cup O|}
\end{equation}

\subsection{Implementation}
\label{sec:implementation}
All components use Qwen3.8-27B in bfloat16 precision.
The modular approach trains a separate LoRA adapter
for each component, while the joint approach trains
one adapter for the complete prediction. We use
LoRA rank 16, scaling factor $\alpha=32$, and no
LoRA dropout.
Models are optimized with AdamW for up to six
epochs on eight NVIDIA B200 GPUs, using a global
batch size of 16. The learning rate is
$2\times10^{-4}$, with 10 warm-up steps followed
by linear decay. Weight decay is 0.001, and
gradients are clipped at a norm of 1.0.
Additional details on the training process are provided in the supplementary.

\section{Results}

\begin{figure*}[t]
\centering
\includegraphics[width=0.8\textwidth]{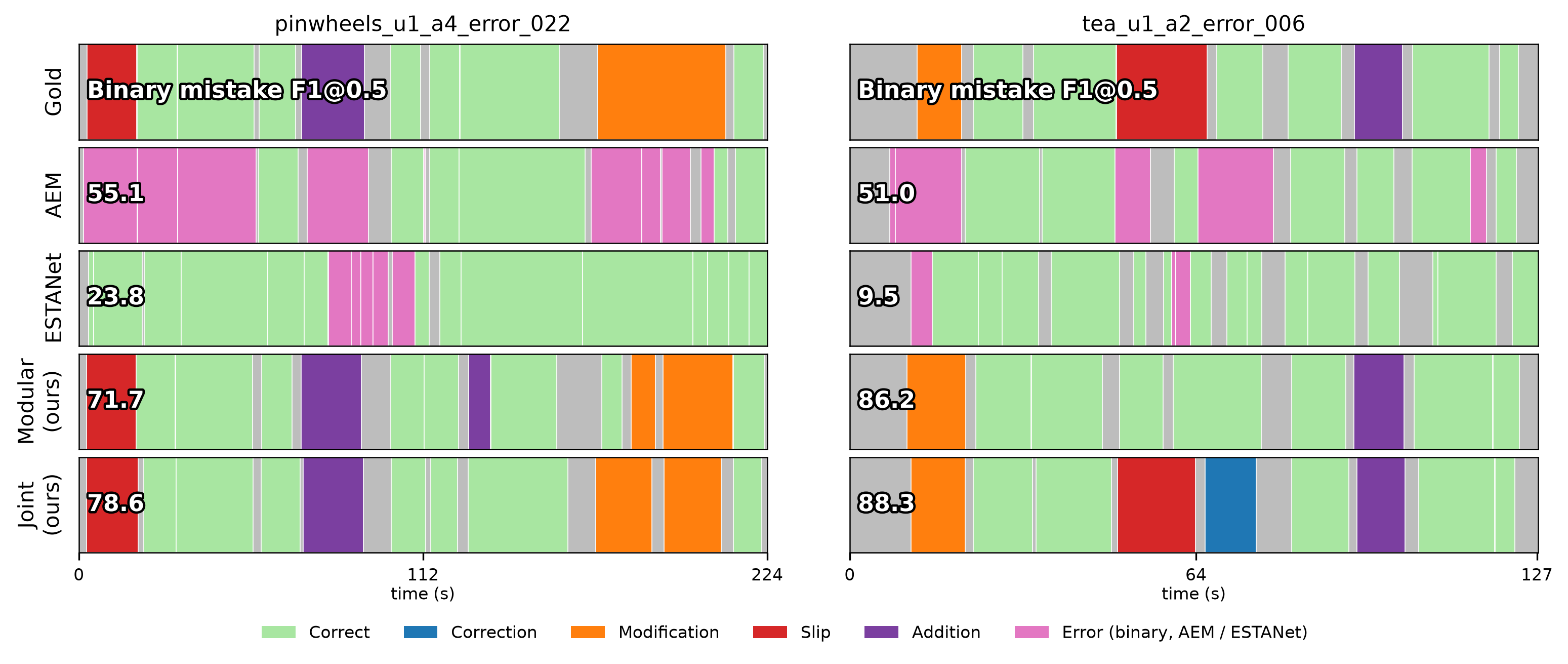}
\caption{Mistake predictions on two test videos. Our approaches assign a class to each segment; AEM and ESTANet give a binary error decision (pink). The number in each row is the binary F1@0.5 on that video. grey marks that no action is detected.}
\label{fig:timeline_mistakes}
\end{figure*}

\begin{table}[t]
\centering
\small
\caption{Error detection (\%). EDA denotes error detection accuracy; F1@0.5 is the macro-average of error and normal F1, with segments matched at temporal IoU~0.5. O-Acc is Omission detection accuracy and O-IoU is omission IoU. Zero-shot uses the untrained backbone with the joint approach's prompt. AEM and ESTANet are evaluated only on the erroneous subset because the remaining test videos overlap with their training or validation data.}
\label{tab:res_detection}
\begin{tabular}{lcccc}
\toprule
Method & EDA & F1@0.5 & O-Acc & O-IoU\\
\midrule
\multicolumn{3}{l}{\emph{Full test set (58 videos)}} \\
Zero-shot & 88.7 & 8.9 & \textbf{100.0} & 8.32 \\
Modular (ours) & 97.7 & 80.7 & 99.6 & 79.1\\
Joint (ours) & \textbf{97.8} & \textbf{84.1} & 97.8 & \textbf{95.4}\\
\midrule
\multicolumn{3}{l}{\emph{Erroneous subset (25 videos)}} \\
Zero-shot & 74.8 & 9.1 & \textbf{100.0} & 16.31\\
AEM~\cite{guo2026aem} & 68.1 & 47.3 & 72.6 & 35.0\\
ESTANet~\cite{lee2026estanet} & 70.0 & 27.0 & 50.4 & 21.3 \\
Modular (ours) & 93.9 & 81.3 & 99.2 & 82.2\\
Joint (ours) & \textbf{94.8} & \textbf{86.2} & 94.9 & \textbf{93.8} \\
\bottomrule
\end{tabular}
\end{table}

\begin{table*}[t]
\centering
\small
\caption{Fine-grained mistake detection: precision (P),
recall (R), and F1 at temporal IoU~0.5 (\%). Macro P, R and F1 are unweighted means over the four mistake types. $n$ denotes the number of annotated test
segments of each type.}
\label{tab:res_fine}
\begin{tabular}{llrccc|ccc}
\toprule
& & & \multicolumn{3}{c}{Modular} &
    \multicolumn{3}{c}{Joint} \\
Test set & Mistake type & $n$ & P & R & F1 & P & R & F1 \\
\midrule
\multirow{5}{*}{All 58}
& Correction   & 5  & 40.0 & 80.0 & 53.3 & 30.0 & 60.0 & 40.0 \\
& Modification & 33 & 92.9 & 78.8 & 85.2 & 87.5 & 84.8 & 86.2 \\
& Slip         & 19 & 46.2 & 31.6 & 37.5 & 55.0 & 57.9 & 56.4 \\
& Addition     & 16 & 62.5 & 93.8 & 75.0 & 65.0 & 81.3 & 72.2 \\
& Macro        & -- & 60.4 & 71.0 & 62.8 & 59.4 & 71.0 & \textbf{63.7} \\
\midrule
\multirow{5}{*}{Erroneous 25}
& Correction   & 5  & 40.0 & 80.0 & 53.3 & 30.0 & 60.0 & 40.0 \\
& Modification & 33 & 92.9 & 78.8 & 85.2 & 87.5 & 84.8 & 86.2 \\
& Slip         & 19 & 50.0 & 31.6 & 38.7 & 61.1 & 57.9 & 59.5 \\
& Addition     & 16 & 75.0 & 93.8 & 83.3 & 92.9 & 81.3 & 86.7 \\
& Macro        & -- & 64.5 & 71.0 & 65.2 & 67.9 & 71.0 & \textbf{68.1} \\
\bottomrule
\end{tabular}
\end{table*}

\begin{table}[t]
\centering
\small
\caption{Segmental step F1@0.5 on the full test set and its erroneous-video subset, in \%. Annotated additions are included as targets. Pred. is the number of predicted segments; the full test set contains 694 annotated segments and the erroneous subset 280 (264 steps and 16 additions). AEM and ESTANet are evaluated only on the 25 erroneous videos.}
\label{tab:res_tas}
\begin{tabular}{lcccc}
\toprule
& \multicolumn{2}{c}{Full test set} & \multicolumn{2}{c}{Erroneous (25 videos)} \\
\cmidrule(lr){2-3}\cmidrule(lr){4-5}
Method & F1@0.5 & Pred. & F1@0.5 & Pred. \\
\midrule
AEM~\cite{guo2026aem} & -- & -- & 61.1 & 368 \\
ESTANet~\cite{lee2026estanet} & -- & -- & 46.2 & 387 \\
Modular (ours) & 92.1 & 696 & 91.5 & 283 \\
Joint (ours) & \textbf{93.6} & 724 & \textbf{94.6} & 284 \\
\bottomrule
\end{tabular}
\end{table}

\begin{table}[t]
\centering
\small
\setlength{\tabcolsep}{3pt}
\caption{Component-level performance. All includes all calls;
Cond. includes calls with correct upstream predictions.
Scores are \%, except boundary error (seconds). $n$ - number of calls scored.}
\label{tab:res_modules}
\begin{tabular}{@{}llrrrr@{}}
\toprule
&& \multicolumn{2}{c}{Test (58)} &
   \multicolumn{2}{c}{Erroneous (25)} \\
\cmidrule(lr){3-4}\cmidrule(lr){5-6}
Component & Metric & All & Cond. & All & Cond. \\
\midrule
\multirow{3}{*}{\shortstack{Step\\Identifier}}
& $n$ & 2{,}679 & -- & 965 & -- \\
& Step macro F1 & 80.3 & -- & 83.8 & -- \\
& Accuracy & 88.9 & -- & 90.1 & -- \\
\midrule
\multirow{4}{*}{\shortstack{Temporal\\Localizer}}
& $n$ & 871 & 656 & 333 & 266 \\
& Temporal F1@0.5 & 85.0 & 96.9 & 86.3 & 96.8 \\
& Mean IoU & 81.5 & 84.5 & 82.1 & 84.7 \\
& Boundary error (s) & 2.26 & 1.42 & 2.10 & 1.47 \\
\midrule
\multirow{6}{*}{\shortstack{Execution\\Verifier}}
& $n$ & 794 & 623 & 302 & 245 \\
& Macro F1 & 66.6 & 73.3 & 67.1 & 72.9 \\
& \texttt{CORRECT} F1 & 87.9 & 92.0 & 88.7 & 90.3 \\
& \texttt{CORRECTION} F1 & 57.1 & 61.5 & 57.1 & 61.5 \\
& \texttt{MODIFICATION} F1 & 83.9 & 85.2 & 83.9 & 85.2 \\
& \texttt{SLIP} F1 & 37.5 & 54.5 & 38.7 & 54.5 \\
\bottomrule
\end{tabular}
\end{table}

We report results on our full test (58 videos) from which 25 are erroneous videos. AEM~\cite{guo2026aem} and ESTANet~\cite{lee2026estanet} are scored only on the 25 erroneous videos. 
Exemplary classifications are shown in the supplementary.

\paragraph{Error Detection}
\cref{tab:res_detection} reports
error detection accuracy (EDA) and binary F1@0.5, the
macro-average of F1 for the error and normal classes.
On the full test set, the joint approach achieves 97.8\%
EDA and 84.1\% F1@0.5, compared with 97.7\% and
80.7\% for the modular approach. The joint model's
unadapted Qwen3.8-27B backbone, prompted to produce
the same outputs but without task-specific fine-tuning,
achieves 88.7\% EDA but only 8.9\% F1@0.5. This gap
shows why EDA alone can be misleading: a model can
score highly by favoring the majority normal class while
detecting few errors.

The distinction is also clear on the 25 erroneous videos.
The joint approach reaches 94.8\% EDA and 86.2\%
F1@0.5, while the modular approach reaches 93.9\%
and 81.3\%. AEM obtains 68.1\% EDA and 47.3\%
F1@0.5, and ESTANet obtains 70.0\% and 27.0\%,
respectively. The unadapted backbone scores 74.8\%
EDA but just 9.1\% F1@0.5. Thus, the joint model's
advantage is not limited to the full test set, where
normal videos are common, it also achieves the strongest
F1 on the subset containing only videos with errors.

\paragraph{Fine-Grained Mistakes}
\cref{tab:res_fine} reports precision, recall, and
F1@0.5 for each mistake type. On all 58 test videos,
the joint approach improves macro F1 slightly by 0.9 points
over the modular approach (63.7\% vs. 62.8\%).
The largest gain is for slips (37.5\% vs. 56.4\%). Modification and Addition F1 are similar (85.2\% vs. 86.2\%; 75.0\% vs. 72.2\%), and correction F1
is relatively low for both models, based on only five test
segments.

The pattern is similar on the 25 videos containing
mistakes, where joint macro F1 is 2.9 points higher
(68.1\% vs.\ 65.2\%). The joint model again improves
slip by a large margin, while performing worse on corrections and remaining close on modifications and additions. \cref{fig:timeline_mistakes} shows an example classification for two tasks, where the gap between our methods and the gold standard is much smaller than that of AEM and ESTANet.

\paragraph{Omissions}
\Cref{tab:res_detection} also reports omission accuracy and omission IoU.
On the 25 erroneous videos, 20 of which contain at least one omission (44 omitted steps in total), the zero-shot approach has the highest O-Acc (100.0) and the joint approach the highest O-IoU.  The modular approach reaches 99.2 O-Acc and 82.2 O-IoU and the joint approach 94.9 and 93.8, while AEM reaches 72.6 and 35.0, and ESTANet 50.4 and 21.3. On the full test split, with 48 omitted steps in 23 videos, the modular approach reaches 99.6 O-Acc and 79.1 O-IoU, and the joint approach 97.8 and 95.4. The modular approach leaves more steps unpredicted and therefore flags more omissions, which manifests as higher O-Acc but lower O-IoU.


\paragraph{Temporal Action Segmentation}
\cref{tab:res_tas} reports temporal segmentation using segmental step
F1@0.5. Background is excluded and addition segments are included
as targets; an addition matches a prediction labeled either
as off-script or with the step identity assigned by the
annotation.
The joint approach achieve slightly higher performance
performance on both test sets: 93.6\% vs. 92.1\% F1@0.5,
respectively, on all 58 videos, and 94.6\% vs. 91.5\% on
the 25 erroneous videos.
Their close performance suggests, despite their different prediction
structures, the two approaches segment procedural steps
comparably well. On the erroneous subset, both also produce
segment counts close to the 280 annotated segments
(283 for the modular approach and 284 for the joint
approach), whereas AEM and ESTANet produce 368 and
387 segments, respectively. Their lower F1 scores
(61.1\% and 46.2\%) are consistent with greater
over-segmentation.\footnote{ESTANet outputs frame-level
labels, we aggregate consecutive frames with the same
predicted step into segments for this comparison; we report the detector with the highest score.}

\paragraph{Module-Level Performance}
\cref{tab:res_modules} evaluates the Step Identifier,
Temporal Localizer, and Execution Verifier both on all
their calls and, where applicable, on calls for which
the preceding component produced the correct output.
For each window, the reference is the first annotated
action that ends within it and has not already been
assigned to an earlier prediction with IoU $\geq 0.5$;
if no such action exists, the reference is \texttt{no step}.
The Step Identifier is scored on whether it identifies
this reference correctly. The Temporal Localizer is
scored on its ability to localize the reference action;
its conditional scores include only calls where the
Step Identifier named that action correctly. The Execution
Verifier is scored on the label of the annotated segment
with which its predicted span overlaps most. 

The conditional scores show how much each component's
performance depends on its input. The Temporal Localizer's
F1@0.5 rises from 85.0\% to 96.9\% on the full test set
when the Step Identifier is correct, while its boundary
error falls from 2.26 to 1.42 MAE in seconds. The Execution
Verifier also improves with correct upstream outputs:
macro F1 rises from 66.6\% to 73.3\%, driven largely by
slip F1, which increases from 37.5\% to 54.5\%. Its
\texttt{CORRECT} and \texttt{MODIFICATION} scores are
already high, whereas \texttt{CORRECTION} remains more
difficult, but also improves from 57.1\% to 61.5\%. The same pattern appears on the erroneous
subset. Overall, localization and classification are
strong once their inputs are correct, but upstream errors
reduce end-to-end performance.

\paragraph{Limitations} Our evaluation has several limitations. First, training and testing on EgoPER's five cooking tasks does not establish generalization to other procedures or settings. Second, our split differs from the official split, and baseline comparisons are limited to 25 erroneous test videos due to data overlap. Moreover, differences in supervision and evaluation protocols prevent a fully controlled comparison. Finally, replay-based online evaluation does not capture live interaction, alert usefulness, or error recovery, which remain future work.

\section{Conclusion}
\label{sec:conclusion}
We propose two approaches for online, instruction-conditioned mistake detection in egocentric procedural video. The modular approach uses three components to identify steps, localize them, and assess their execution, while the joint approach performs all three tasks in a single model. Both achieve strong temporal segmentation, and the joint approach detects mistakes more accurately, particularly slips. These findings demonstrate the potential of both approaches for detecting deviations as a procedure unfolds and motivate evaluation on a broader range of tasks and in online settings.

{\small
\bibliographystyle{ieeenat_fullname}
\bibliography{references}

@inproceedings{guo2026aem,
  title={Procedural mistake detection via action effect modeling},
  author={Guo, Wenliang and Pu, Yujiang and Kong, Yu},
  booktitle={International Conference on Learning Representations},
  volume={2026},
  pages={43009--43030},
  year={2026}
}

@article{bacharidis2025review,
  title={Vision-based mistake analysis in procedural activities: A review of advances and challenges},
  author={Bacharidis, Konstantinos and Argyros, Antonis},
  journal={Computer Vision and Image Understanding},
  pages={104842},
  year={2026},
  publisher={Elsevier}
}

@inproceedings{haneji2024egooops,
  title={{EgoOops}: A dataset for mistake action detection from egocentric videos referring to procedural texts},
  author={Haneji, Yuto and Nishimura, Taichi and Kameko, Hirotaka and Shirai, Keisuke and Yoshida, Tomoya and Kajimura, Keiya and Yamamoto, Koki and Cui, Taiyu and Nishimoto, Tomohiro and Mori, Shinsuke},
  booktitle={2025 IEEE/CVF International Conference on Computer Vision Workshops (ICCVW)},
  pages={2711--2721},
  year={2025},
  organization={IEEE}
}

@inproceedings{lee2024egoper,
  title={Error detection in egocentric procedural task videos},
  author={Lee, Shih-Po and Lu, Zijia and Zhang, Zekun and Hoai, Minh and Elhamifar, Ehsan},
  booktitle={Proceedings of the IEEE/CVF Conference on Computer Vision and Pattern Recognition},
  pages={18655--18666},
  year={2024}
}

@inproceedings{flaborea2024prego,
  title={{PREGO}: online mistake detection in procedural egocentric videos},
  author={Flaborea, Alessandro and Di Melendugno, Guido Maria D'Amely and Plini, Leonardo and Scofano, Luca and De Matteis, Edoardo and Furnari, Antonino and Farinella, Giovanni Maria and Galasso, Fabio},
  booktitle={2024 IEEE/CVF Conference on Computer Vision and Pattern Recognition (CVPR)},
  pages={18483--18492},
  year={2024},
  organization={IEEE}
}

@inproceedings{sener2022assembly101,
  title={Assembly101: A large-scale multi-view video dataset for understanding procedural activities},
  author={Sener, Fadime and Chatterjee, Dibyadip and Shelepov, Daniel and He, Kun and Singhania, Dipika and Wang, Robert and Yao, Angela},
  booktitle={2022 IEEE/CVF Conference on Computer Vision and Pattern Recognition (CVPR)},
  pages={21064--21074},
  year={2022},
  organization={IEEE}
}

@article{hu2022lora,
  title={Lora: Low-rank adaptation of large language models},
  author={Hu, Edward J and Shen, Yelong and Wallis, Phillip and Allen-Zhu, Zeyuan and Li, Yuanzhi and Wang, Shean and Wang, Lu and Chen, Weizhu},
  journal={arXiv preprint arXiv:2106.09685},
  year={2021}
}

@inproceedings{jang2019epic,
  title={{EPIC-Tent}: An egocentric video dataset for camping tent assembly},
  author={Jang, Youngkyoon and Sullivan, Brian and Ludwig, Casimir and Gilchrist, Iain D and Damen, Dima and Mayol-Cuevas, Walterio},
  booktitle={2019 IEEE/CVF International Conference on Computer Vision Workshop (ICCVW)},
  pages={4461--4469},
  year={2019},
  organization={IEEE}
}

@inproceedings{huang2025amnar,
  title={Modeling multiple normal action representations for error detection in procedural tasks},
  author={Huang, Wei-Jin and Li, Yuan-Ming and Xia, Zhi-Wei and Tang, Yu-Ming and Lin, Kun-Yu and Hu, Jian-Fang and Zheng, Wei-Shi},
  booktitle={2025 IEEE/CVF Conference on Computer Vision and Pattern Recognition (CVPR)},
  pages={27794--27804},
  year={2025},
  organization={IEEE}
}

@inproceedings{lee2025gtg2vid,
  title={Error recognition in procedural videos using generalized task graph},
  author={Lee, Shih-Po and Elhamifar, Ehsan},
  booktitle={2025 IEEE/CVF International Conference on Computer Vision (ICCV)},
  pages={10009--10021},
  year={2025},
  organization={IEEE}
}

@article{ozsoy2026zeprom,
  title={The Unreasonable Effectiveness of VLMs for Zero-shot Procedural Mistake Detection},
  author={Ozsoy, Serdar and Doorenbos, Lars and Spurio, Federico and Francesca, Gianpiero and Gall, Juergen},
  journal={arXiv preprint arXiv:2606.21579},
  year={2026}
}

@inproceedings{lee2026estanet,
  title={{ESTANet}: Efficient Online Error Detection in Procedural Videos via Prediction Inconsistency},
  author={Lee, Shih-Po and Ghoddoosian, Reza and Siddiqui, Faizan and Sachdeva, Enna and Dariush, Behzad},
  booktitle={European Conference on Computer Vision},
  pages={117--134},
  year={2026},
  organization={Springer}
}

@article{wen2026provia,
  title={PROVIA: Procedure State Tracking for Online Mistake Detection in Egocentric Videos},
  author={Wen, Di and Yang, Kailun and Weissert, Jimmy and Scherrer, Luc Maria and Z{\"o}llner, Cedric and Liu, Ruiping and Chen, Yufan and Wei, Jiale and Zheng, Junwei and Peng, Kunyu},
  journal={arXiv preprint arXiv:2609.20638},
  year={2026}
}

@article{zhang2019bertscore,
  title={Bertscore: Evaluating text generation with bert},
  author={Zhang, Tianyi and Kishore, Varsha and Wu, Felix and Weinberger, Kilian Q and Artzi, Yoav},
  journal={arXiv preprint arXiv:1904.09675},
  year={2019}
}
}


\clearpage
\appendix
\clearpage
\setcounter{page}{1}
\maketitlesupplementary

\section{Example Performance}
\label{app:example_image}
\newcommand{\exfield}[2]{\textbf{#1:} #2\\}
\newcommand{\exboxheight}{66pt} 

\begin{figure*}[t]
\centering
\begin{minipage}[t]{0.485\textwidth}
\centering
\includegraphics[width=\linewidth]{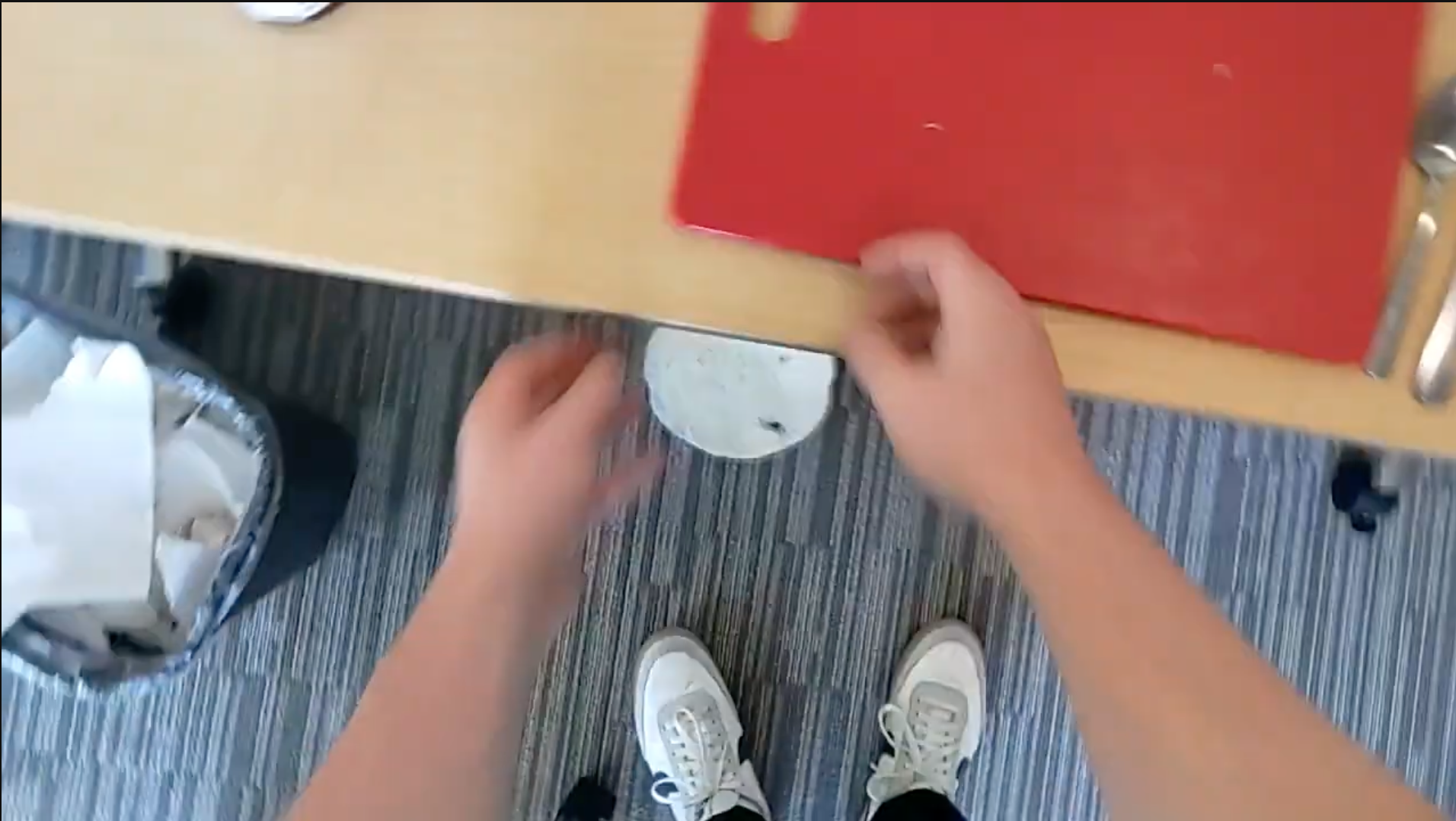}\\[3pt]
\fbox{\parbox[t][\exboxheight][t]{\dimexpr\linewidth-2\fboxsep-2\fboxrule\relax}{\small\raggedright
\exfield{Approach}{Joint}
\exfield{Instruction}{Place tortilla on cutting board}
\exfield{Span}{2.5--19.4\,s (decided at 20.0\,s)}
\exfield{Class}{Slip}
\textbf{Reasoning:} ``Drop tortilla on floor''}}
\\[2pt]{\small (a)}
\end{minipage}\hfill
\begin{minipage}[t]{0.485\textwidth}
\centering
\includegraphics[width=\linewidth]{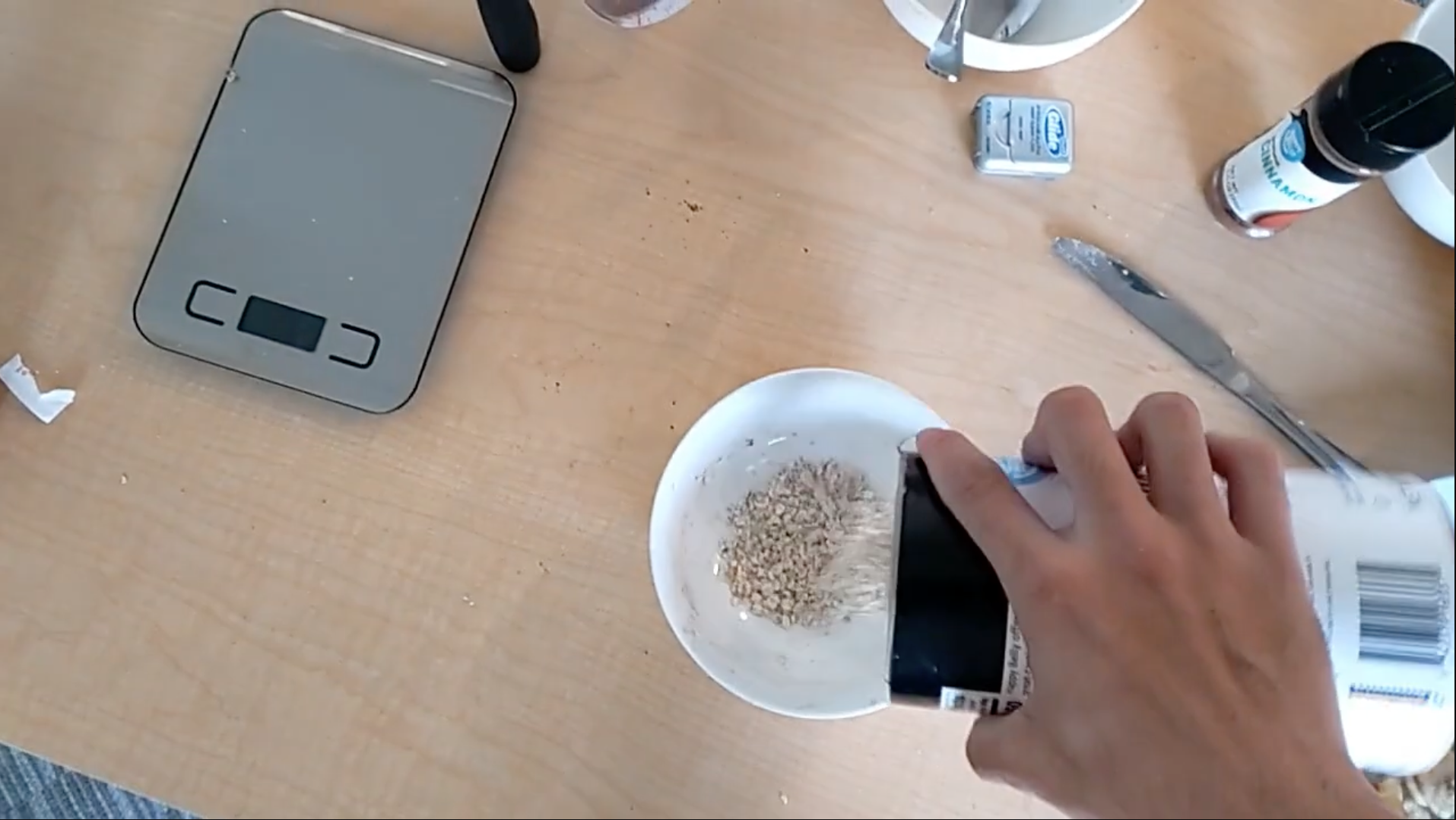}\\[3pt]
\fbox{\parbox[t][\exboxheight][t]{\dimexpr\linewidth-2\fboxsep-2\fboxrule\relax}{\small\raggedright
\exfield{Approach}{Modular}
\exfield{Instruction}{Measure 4 Tablespoons of quick-cook oats and put them into bowl}
\exfield{Span}{2.5--13.8\,s (decided at 15.0\,s)}
\exfield{Class}{Modification}
\textbf{Reasoning:} ``Directly pour quick oats into bowl''}}
\\[2pt]{\small (b)}
\end{minipage}
\caption{Examples of detected mistakes. (a) The joint approach flags a slip: the tortilla is dropped on the floor instead of being placed on the cutting board. (b) The modular approach flags a modification: the oats are poured directly from the container instead of being measured with a tablespoon. Span: predicted start and end of the action; decided at: the video time at which the decision was made.}
\label{fig:examples}
\end{figure*}

\cref{fig:examples} present images of correctly detected mistakes by the models. This is also a demonstration of the output of each of the models that contain the instruction of the action, the time span, the classification, and the reasoning behind the classification.

\begin{figure*}[t]
\centering
\includegraphics[width=\textwidth]{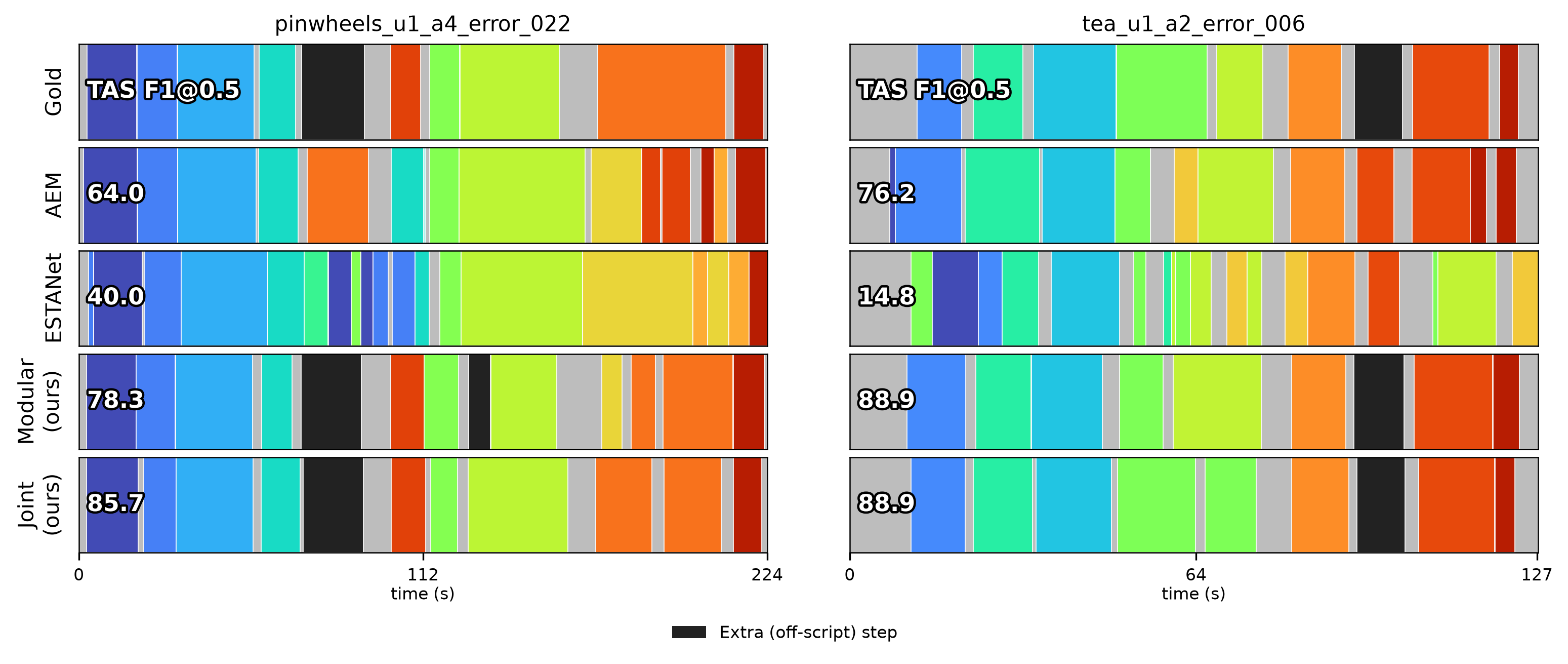}
\caption{Step segmentation on two test videos. Each color is one step of the recipe; grey is background and black an off-script step. The number in each row is the segmental F1@0.5 on that video. }
\label{fig:timeline_steps}
\end{figure*}

\cref{fig:timeline_steps,fig:timeline_mistakes} compare the step segmentation and the mistake predictions of both our methods, AEM and ESTANet on two erroneous test videos from different tasks, pinwheels and tea. On step segmentation (\cref{fig:timeline_steps}), the joint approach reaches a segmental F1@0.5 of 85.7 on pinwheels and 88.9 on tea, the modular approach 78.3 and 88.9, AEM 64.0 and 76.2, and ESTANet 40.0 and 14.8. Both of our approaches mark the annotated off-script step in each video. On mistake detection (\cref{fig:timeline_mistakes}), the binary F1@0.5, averaged over the error and the normal class, is 78.6 and 88.3 for the joint approach, 71.7 and 86.2 for the modular approach, 55.1 and 51.0 for AEM, and 23.8 and 9.5 for ESTANet. AEM and ESTANet give a binary error decision, whereas our approaches also assign a mistake type. On pinwheels, both of our approaches approaches detect the slip and the addition; the modification is split into two segments, neither of which reaches IoU 0.5, so it counts as missed. On tea, both of our approaches detect the modification and the addition; the joint approach also detect the slip.

\section{Implementation Details}
\label{sec:implementation}

\paragraph{Shared setup.}
All four models (Modules A, B and C of the modular approach, and the joint approach) are fine-tuned separately from the same backbone, Qwen3.8-27B, in bf16. Each model is a LoRA adapter~\cite{hu2022lora} of rank 16, with $\alpha=32$ and no dropout; the backbone weights stay frozen, and no layer is trained in full. All four adapters place LoRA on the same modules of the language model (\cref{tab:lora}). The 64 decoder layers are of two kinds, linear-attention (Gated DeltaNet) layers and full-attention layers, and each kind gets its own set of targets; the MLP of every layer is adapted as well. The adapters differ only in the vision encoder (\cref{tab:impl}).

\begin{table}[ht]
\centering
\footnotesize
\setlength{\tabcolsep}{3pt}
\caption{LoRA targets in the language model, shared by all four adapters.}
\label{tab:lora}
\begin{tabular}{lcl}
\toprule
Block & Layers & LoRA targets \\
\midrule
Linear attention & 48 & \makecell[l]{\texttt{in\_proj\_qkv}, \texttt{in\_proj\_z}, \texttt{in\_proj\_a},\\ \texttt{in\_proj\_b}, \texttt{out\_proj}} \\
Full attention & 16 & \texttt{q\_proj}, \texttt{k\_proj}, \texttt{v\_proj}, \texttt{o\_proj} \\
MLP & 64 & \texttt{gate\_proj}, \texttt{up\_proj}, \texttt{down\_proj} \\
\bottomrule
\end{tabular}
\end{table}

We train with fused AdamW at a learning rate of $2\times10^{-4}$, 10 warm-up steps, linear decay, weight decay 0.001 and gradient clipping at 1.0, for up to six epochs on eight NVIDIA B200 GPUs. Each GPU holds one example and gradients are accumulated over two steps, giving a global batch of 16. Video is decoded at 1 frame per second. During inference the stream is polled every 5\,s of video time. \cref{tab:impl} lists the settings that differ between the models.

\begin{table}[t]
\centering
\footnotesize
\setlength{\tabcolsep}{3pt}
\caption{Per-model settings. Vision: whether LoRA adapters are also placed in the vision encoder (adapted) or the encoder is left unchanged (frozen). Frames: max number of video frames in one input. Seq.: max input length in tokens. Gen.: max number of generated tokens. Epoch: checkpoint used at test time.}
\label{tab:impl}
\resizebox{\columnwidth}{!}{%
\begin{tabular}{lccccc}
\toprule
Model & Vision & Frames & Seq. & Gen. & Epoch \\
\midrule
Step Identifier & adapted & 32 & 12{,}288 & 1 letter & 6 \\
Temporal Localizer & frozen & 54 & 20{,}480 & 64 & 4 \\
Execution Verifier & frozen & 40 & 20{,}480 & 128 & 4 \\
Joint & adapted & 54 & 20{,}480 & 192 & 6 \\
\bottomrule
\end{tabular}%
}
\end{table}

\paragraph{Parameters.}
\begin{itemize}
\item \emph{LoRA rank and $\alpha$.} Each adapted weight matrix $W$ is replaced by $W+\frac{\alpha}{r}BA$, where $A$ and $B$ are trainable low-rank matrices of rank $r$. With $r=16$ and $\alpha=32$ the update is scaled by 2.
\item \emph{Vision adapted / frozen.} For Module A and the joint approach, LoRA adapters are also placed in the vision encoder, so the visual features are fine-tuned. For Modules B and C, only the language model is adapted.
\item \emph{Frames.} The largest number of frames passed in one input. A clip longer than this cap at 1\,fps is subsampled uniformly over its whole duration, from its start to its end.
\item \emph{Sequence length.} The largest number of tokens in one training example: system and user prompt, video tokens and target.
\item \emph{Generation budget.} The largest number of tokens the model may produce. Module A's answer is a single letter chosen by scoring the candidate letters, not by free generation.
\item \emph{Checkpoint selection.} The test checkpoint of each model is the one with the best validation score: step macro F1 for Module A, temporal F1@0.5 for Module B, and class macro F1 for Module C. For the joint approach it is the geometric mean of these three metrics.

\end{itemize}

\newcommand{\promptbox}[2]{%
\par\noindent\fbox{\parbox{\dimexpr\columnwidth-2\fboxsep-2\fboxrule\relax}{%
\scriptsize\raggedright\textbf{#1}\par\smallskip\ttfamily #2}}\par\smallskip}
\newcommand{\expected}[1]{%
\noindent{\scriptsize\raggedright\textbf{Expected output.} #1\par}\medskip}

\paragraph{Prompt examples.}
The prompts below are shown for the oatmeal recipe. Each system prompt is fixed; the user prompt is filled in for each call. The step lists are shortened here to the first two steps; the actual prompts list every step of the recipe.

\promptbox{Module A --- system}{%
Detect which procedural step, if any, completed in the clip. Reply with exactly one choice letter. A means no step completed. B means an extra off-instruction step completed. C and later letters identify a listed instruction that just finished. If the person did a listed action incorrectly, still pick that listed letter when it is the closest match (for example, measuring 13 ounces still maps to Measure 12 ounces of cold water using measuring cup). Use B only when the completed action is not any listed instruction. Do not assume recipe order.}

\promptbox{Module A --- user}{%
Task: oatmeal\\
Task description: Prepare oatmeal.\\
Dataset: egoper\\
Completed instructions before this clip:\\
- None\\
Choices:\\
A: NONE (no newly completed procedural event)\\
B: ADDITION (an off-instruction procedural event completed)\\
C: STEP\_0 {\rmfamily ---} Measure 12 ounces of cold water using measuring cup\\
D: STEP\_1 {\rmfamily ---} Pour the water into the bowl containing oats\\
\ldots\ (one letter for every remaining step of the recipe)}
\expected{If measuring the water has finished, even if it was done incorrectly (e.g.\ 13 ounces instead of 12): \texttt{C}. If the measuring is still underway: \texttt{A}.}

\promptbox{Module B --- system}{%
Locate the requested instruction attempt in the clip. If it is completed, return JSON only as \{"start":<seconds>,"end":<seconds>\} using seconds relative to the start of the clip. If the attempt is not completed, return JSON only as \{"status":"NOT\_COMPLETED"\}.}

\promptbox{Module B --- user}{%
Instruction: Measure 12 ounces of cold water using measuring cup\\
The clip-local physical prediction region is [0.00, 18.40] seconds. A valid span must satisfy 0.00 <= start < end <= 18.40.}
\expected{If the attempt is complete in the clip: \texttt{\{"start":4.20, "end":16.80\}}. If Module A fired too early and the attempt is not finished: \texttt{\{"status":"NOT\_COMPLETED"\}}.}

\promptbox{Module C --- system}{%
Classify the visible execution of the requested instruction. Choose exactly one outcome. CORRECTION is a repair of a preceding error and is not itself a mistake. If the clip does not contain a completed attempt, return \{"status":"NOT\_COMPLETED"\}. Return JSON only as \{"classification":"<class>","reasoning":"<visual evidence>"\} or \{"status":"NOT\_COMPLETED"\}.}

\promptbox{Module C --- user}{%
Dataset: egoper\\
Instruction: Pour the water into the bowl containing oats\\
Classify only the execution visible in this bounded clip. Allowed outcomes: CORRECT, CORRECTION, MODIFICATION, SLIP, NOT\_COMPLETED. SLIP means executing the requested step in a way that fails to achieve its goal, such as adding water to a different bowl from the one containing oats or dropping a tortilla on the floor. MODIFICATION means performing the requested step differently from the recipe, such as scooping nut butter with a spoon or pouring water without the specified circular motion. This does not necessarily change the step outcome. CORRECTION means performing an action that mitigates the effect of a preceding SLIP, such as discarding a tortilla dropped on the floor and picking up a new one. CORRECT means the visible execution follows the instruction. NOT\_COMPLETED means this clip is not a completed attempt. Describe the observable evidence without repeating the class name as the explanation.\\
Recent prior action history (up to the 12 most recent eligible completed actions; shown oldest to newest; older actions may be omitted; use only when relevant):\\
\ \ 1. Action: Measure 12 ounces of cold water using measuring cup | Classification: SLIP | Evidence: Water was poured into an empty bowl instead of the bowl containing oats.}
\expected{\texttt{\{"classification":"CORRECTION", "reasoning":"The water is transferred from the empty bowl into the bowl containing oats."\}}}

\promptbox{Joint approach --- system}{%
Review the accumulated causal video once. Decide whether a procedural action has finished. If none has finished, return exactly \{"status":"NOT\_COMPLETED"\}. Otherwise return exactly one JSON object with keys "step\_id","start","end","classification","reasoning". step\_id must be STEP\_k for a listed instruction or ADDITION for an off-script action. Times are seconds relative to the beginning of this clip and must satisfy 0 <= start < end <= the stated clip duration. CORRECTION repairs a preceding error and is not itself a mistake. Return JSON only.}

\promptbox{Joint approach --- user}{%
Dataset: EgoPER\\
Task: oatmeal\\
Task description: Prepare oatmeal.\\
Clip-local prediction region: [0.00, 18.40] seconds.\\
Procedure:\\
\ \ STEP\_0: Measure 12 ounces of cold water using measuring cup\\
\ \ STEP\_1: Pour the water into the bowl containing oats\\
\ \ \ldots\ (every remaining step of the recipe)\\
Predicted prior eligible actions (up to 12, oldest to newest):\\
\ \ (none)\\
Class definitions: CORRECT follows the listed instruction; SLIP performs it but fails its goal; MODIFICATION performs it differently from the procedure; CORRECTION repairs an earlier deviation; ADDITION is a completed off-script action. Determine only what completed in this causal clip.}
\expected{If nothing has finished: \texttt{\{"status":"NOT\_COMPLETED"\}}. If measuring the water finished with a slip: \texttt{\{"step\_id":"STEP\_0", "start":4.20, "end":16.80, "classification":"SLIP", "reasoning":"Water was poured into an empty bowl instead of the bowl containing oats."\}}.}

\section{Additional Results}
\label{sec:app_results}

\paragraph{Explanation quality.}
\cref{tab:app_expl} compares the reasoning each approach writes for a correctly detected mistake with the reference explanation. A detected mistake is a predicted mistake segment that overlaps an annotated mistake segment with IoU of at least 0.5. N is the number of these segments that have reasoning text. The joint approach writes reasoning for every one of them (59). The modular approach detects 56, but 15 are additions, which skip Module C and therefore have no reasoning, leaving 41. \emph{Exact match (EM)} is the percentage of explanations that are identical to the reference text. The remaining three columns are BERTScore F1~\cite{zhang2019bertscore} scores. \emph{Own reference} compares each explanation with its own reference. We add two controls by scoring texts that are related but are not the explanation of the mistake: \emph{Instruction} compares the reference with the step's instruction, and \emph{Wrong reference} compares each explanation with the reference of another mistake from the same recipe.

\begin{table}[t]
\centering
\small
\setlength{\tabcolsep}{4pt}
\caption{Explanations on correctly detected mistakes, full test split. BERTScore F1 with \texttt{deberta-xlarge-mnli} layer~40, no IDF weighting and no baseline rescaling. N: detected mistakes with reasoning text. EM: exact match, the percentage of explanations identical to the reference text.}
\label{tab:app_expl}
\resizebox{\columnwidth}{!}{%
\begin{tabular}{lccccc}
\toprule
& & & \multicolumn{3}{c}{BERTScore F1 against} \\
\cmidrule(lr){4-6}
Method & N & EM (\%) & \makecell{\textbf{Own}\\\textbf{reference}} & \makecell{Instruc-\\tion} & \makecell{Wrong\\reference} \\
\midrule
Modular (ours) & 41 & 75.6 & \textbf{0.899} & 0.730 & 0.631 \\
Joint (ours) & 59 & 86.4 & \textbf{0.944} & 0.685 & 0.643 \\
\bottomrule
\end{tabular}%
}
\end{table}

\paragraph{Latency.}
\cref{tab:app_latency} reports model-call durations on one NVIDIA B200 GPU per video, on the 58 test videos. They exclude data loading and start-up. In the modular approach, Module B is called only after Module A names a step, and Module C only after Module B places a span, so B and C are called less often than A. One accepted action is the sum of the calls that produce one placed action: A, B and C for the modular approach, and a single call for the joint approach. The calls run one after another, and the models run in bf16 without quantization or other inference optimization; we made no attempt to reduce latency, so these times are not a deployed real-time budget.

\begin{table}[t]
\centering
\small
\caption{Model-call durations in seconds on the full test split.}
\label{tab:app_latency}
\resizebox{\columnwidth}{!}{%
\begin{tabular}{llccc}
\toprule
Method & Measurement & Mean & Median & P95 \\
\midrule
Modular & Module A, one call & 1.75 & 1.49 & 3.06 \\
 & Module B, one call & 3.09 & 2.78 & 5.17 \\
 & Module C, one call & 3.10 & 2.95 & 4.53 \\
 & One accepted action & 7.63 & 7.18 & 11.81 \\
\midrule
Joint & One call (all) & 3.18 & 2.22 & 6.42 \\
 & Window with no action & 2.29 & 1.72 & 4.78 \\
 & One accepted action & 5.58 & 5.36 & 7.89 \\
\bottomrule
\end{tabular}%
}
\end{table}

\begin{table}[t]
\centering
\small
\setlength{\tabcolsep}{4pt}
\caption{Scores from ESTANet's evaluator on the 25 erroneous videos, in \%.}
\label{tab:app_estanet}
\begin{tabular}{lcccc}
\toprule
Method & F1@0.1 & F1@0.25 & F1@0.5 & Frame acc. \\
\midrule
AEM~\cite{guo2026aem} & 53.9 & 50.1 & 34.0 & 74.0 \\
ESTANet~\cite{lee2026estanet} & 46.5 & 35.5 & 18.8 & 72.7 \\
Modular (ours) & 79.2 & 78.8 & 72.5 & 92.8 \\
Joint (ours) & 82.8 & 81.7 & 79.2 & 93.9 \\
\bottomrule
\end{tabular}
\end{table}

\paragraph{ESTANet's evaluation protocol.}
ESTANet~\cite{lee2026estanet} does not predict segments. It labels each frame of the video as error or normal, and its evaluator scores this frame-by-frame timeline. The timeline is split into runs of consecutive frames with the same label, and predicted runs are matched one to one with annotated runs of the same label at IoU thresholds of 0.1, 0.25 and 0.5 (F1@0.1, F1@0.25, F1@0.5). Both error and normal runs are counted, and counts are pooled per recipe before averaging over recipes. Frame acc. is the fraction of frames with the correct label. Our segment F1@0.5 (\cref{tab:res_detection}) uses the same threshold, but the annotation it is scored against is different. Our reference is the list of annotated step segments: every step is its own target, with its own start and end, and two neighbouring correct steps remain two separate normal targets. ESTANet's reference is a frame-by-frame error/normal label derived from the same annotation. Neighbouring frames with the same label form one run, so consecutive correct steps merge into a single normal run and the boundaries between them disappear. A long normal prediction that spans several correct steps therefore matches one run under ESTANet's evaluator but none of the individual steps under ours. \cref{tab:app_estanet} scores all methods with ESTANet's released evaluator on the 25 erroneous videos. ESTANet's evaluator takes an error or normal label for every frame. For AEM and our approaches, we mark a frame as error when it lies inside a predicted mistake segment, and as normal otherwise. 
As a reproducibility check, the released ESTANet checkpoints reproduce its published segmental F1 scores on the full official test set within 0.6 percentage points. Under AEM's own threshold-averaged evaluation the checkpoint reproduces its published results on the full official test.

\end{document}